# Rethinking Multimodal Sentiment Analysis: A High-Accuracy, Simplified Fusion Architecture

Nischal Mandal, Independent Researcher, London, UK, Email: nishchalmandal@gmail.com,
Dr. Yang Li, Senior Lecturer, University of East London, UK, Email: Y.Li@uel.ac.uk

*Abstract*— Multimodal sentiment analysis, a pivotal task in affective computing, seeks to understand human emotions by integrating cues from language, audio, and visual signals. While many recent approaches leverage complex attention mechanisms and hierarchical architectures, we propose a lightweight, yet effective fusion-based deep learning model tailored for utterance-level emotion classification. Using the benchmark IEMOCAP dataset, which includes aligned text, audio-derived numeric features, and visual descriptors, we design a modality-specific encoder using fully connected layers followed by dropout regularization. The modality-specific representations are then fused using simple concatenation and passed through a dense fusion layer to capture cross-modal interactions. This streamlined architecture avoids computational overhead while preserving performance, achieving a classification accuracy of 92% across six emotion categories. Our approach demonstrates that with careful feature engineering and modular design, simpler fusion strategies can outperform or match more complex models, particularly in resource-constrained environments.

## 1. INTRODUCTION

Sentiment analysis — the computational task of identifying emotions, attitudes, or opinions from text — has long served as a foundational pillar in the field of natural language processing (NLP). With the rapid growth of multimodal data encompassing speech, facial expressions, gestures, and visual cues, traditional text-centric approaches are increasingly being supplemented by models that capture non-verbal emotional signals. This has led to the emergence of multimodal sentiment analysis (MSA), which aims to fuse heterogeneous modalities to better interpret and classify human emotions.

Research in this domain has demonstrated that sentiment is rarely expressed through a single channel. For instance, sarcasm may appear as positive in text but is revealed as negative through tone and facial expression. Recognizing this, modern MSA frameworks seek to integrate textual semantics, acoustic prosody, and visual behaviour to more effectively model human affect. Studies such as those in [1] have shown the potential of MSA in improving classification accuracy, particularly in emotionally ambiguous contexts.
To support this evolution, several benchmark datasets have been developed — chief among them the IEMOCAP (Interactive Emotional Dyadic Motion Capture) dataset [2]. This corpus contains approximately 12 hours of multimodal data collected from actors engaged in dyadic conversations, annotated with both categorical and dimensional emotion labels. It offers aligned transcripts, audio recordings, and video frames capturing facial expressions and gestures. Designed to emulate real-life scenarios, IEMOCAP addresses critical limitations in earlier corpora, such as lack of spontaneity, limited emotion diversity, or absence of multimodal synchronization.

However, with greater modality richness comes increased model complexity. Recent advances in MSA have focused heavily on deep neural networks, particularly attention-based transformers and hierarchical spatio-temporal models [3], [4]. These architectures are powerful in capturing intra- and inter-modal dependencies, but they often introduce significant computational overhead, making them less suitable for real-time applications or deployment on mobile or edge devices [5]. Moreover, many of these models suffer from overfitting in low-resource settings and offer limited interpretability.

Considering these limitations, we revisit the design of multimodal architectures with an emphasis on simplicity, modularity, and efficiency. We argue that it is possible to achieve competitive even state-of-the-art performance without resorting to overengineered solutions. Our approach is inspired by Occam's Razor: when simpler methods suffice, complexity should be avoided.

### 1.1 Our Contribution

In this paper, we introduce a streamlined, fusion-based deep learning model for multimodal sentiment classification. Each modality is encoded separately using fully connected dense layers with dropout regularization, avoiding convolutional or recurrent components. The encoded features are then fused via a straightforward concatenation mechanism, followed by a joint representation layer and a softmax classifier.

We evaluate our method on the IEMOCAP dataset and achieve 92% classification accuracy across six emotional categories. This performance is on par with and in some cases exceeds the accuracy reported by more elaborate architectures such as Multimodal Transformers [6] and Memory Fusion Networks [3], while requiring fewer parameters, shorter training time, and lower computational resources.

### 1.2 Why IEMOCAP?

The choice of the IEMOCAP dataset is deliberate. It is one of the most widely adopted benchmarks in multimodal sentiment analysis and provides comprehensive annotations across multiple modalities. Each recording session includes:

- Audio captured at 48 kHz using professional-grade microphones;
- Video captured using high-resolution digital cameras, along with marker-based motion capture of facial and hand movements;
- Textual transcripts manually transcribed and aligned with speech;
- Emotion annotations using both categorical (e.g., happy, sad, frustrated) and dimensional (valence, arousal, dominance) frameworks.

The dataset's diverse emotional scope and precise multimodal synchronization make it particularly suitable for evaluating the generalizability and scalability of MSA models.

**1.3 Architectural Simplicity as a Strength**
While state-of-the-art MSA systems frequently involve hierarchical recurrent units, cross-modal attention blocks, or gated memory cells, our model deliberately avoids such complexity. Instead, it focuses on the modality-wise transformation of raw features using dense encoders. The three encoded streams — from text (TF-IDF vectors), audio (handcrafted statistical features), and video (pre-extracted geometric and motion features) — are unified via concatenation.
This late fusion strategy, although simplistic in nature, is highly interpretable and less prone to overfitting. Moreover, our modular design supports independent tuning and replacement of encoders, making the framework extensible to other modalities or datasets. It also ensures that the model is computationally light, enabling training on single-GPU setups and deployment on resource-constrained systems.

**1.4 Empirical Validation**
Our model is extensively benchmarked against early fusion, late fusion, and attention-based fusion baselines. We find that while early fusion strategies struggle with modality alignment and noise, attention-based models offer only marginal gains at significant computational cost. In contrast, our hierarchical dense-layer fusion approach consistently achieves superior results in terms of accuracy, F1-score, and inference latency.
We further validate our findings by analyzing performance across minority emotion classes (e.g., fear, excitement), demonstrating that our oversampling strategy and regularized architecture offer balanced generalization across class distributions. The final model achieves:

- 92.5% test accuracy
- 92.3% weighted precision
- 92.4% weighted recall
- Low false positive rates across minority classes

## 2. RELATED WORK
**2.1 Core Sentiment Analysis Techniques**

Sentiment analysis has been central to NLP research, particularly for opinion mining from textual content. Early efforts by Pang and Lee applied rule-based and statistical learning approaches to identify sentiment polarity [7]. With the rise of deep learning, models began capturing hierarchical linguistic representations, enabling more accurate classification across domains. Cambria et al. highlighted this evolution, noting how neural architectures improved sentiment inference across diverse datasets [8]. Transfer learning further advanced the field, allowing models like DANN and mSDA to adapt to new domains with minimal labelled data [9].

Transformers, especially BERT, revolutionized text sentiment classification by modeling bidirectional context [10]. However, even these powerful models are limited in capturing emotional subtleties without multimodal cues, especially in sarcasm or irony.

**2.2 Multimodal Sentiment Analysis**
The challenge of detecting sentiment from speech and video has led to the rise of multimodal sentiment analysis (MSA), which integrates text, audio, and visual cues. Early approaches relied on feature concatenation, or early fusion, which lacks the capacity to model complex cross-modal dependencies. Later, methods like late fusion attempted to combine outputs from modality-specific models but often underperformed due to loss of inter-modal dynamics [11].
The Tensor Fusion Network (TFN) by Zadeh et al. was a breakthrough that explicitly modeled unimodal, bimodal, and trimodal interactions through a tensor-based framework [12]. TFN demonstrated superior performance but at the cost of high memory consumption. Building on this, the Multimodal Transformer (MMT) introduced cross-modal attention to align unaligned sequences across modalities [13]. These advances showcased how inter-modal attention improves performance but introduced significant computational overhead.
Researchers also explored adversarial learning and modality-to-modality translation to improve robustness and learn shared representations [18], [19]. For example, Mai et al. used graph-based fusion networks to translate features between modalities and improve alignment, while M-SENA offered a unified platform for simplified end-to-end modeling [20].

**2.3 Text-Based Sentiment Modeling**
Textual analysis remains a critical component of MSA. NLP pipelines commonly employ lemmatization, stopword removal, and vectorization using TF-IDF or embeddings like Word2Vec. Libraries such as Stanford CoreNLP [14] and spaCy [15] have enabled efficient preprocessing.
Beyond BERT [10], transformer variants like RoBERTa and ALBERT have been proposed for better generalization. However, their limitations persist when used in isolation. Studies indicate that purely text-based sentiment classifiers often misclassify emotionally ambiguous expressions that require acoustic or visual disambiguation.

**2.4 Audio-Based Emotion Recognition**
Prosodic features such as pitch, rhythm, intensity, and silence play a major role in sentiment perception. Traditional approaches use mel-frequency cepstral coefficients (MFCCs), root mean square energy, and zero-crossing rate. More advanced models have adopted CNNs and transformers for spectrogram-based audio processing [16].

Hybrid methods combining acoustic and linguistic inputs have shown effectiveness, particularly the early work by Schuller et al., which fused MFCCs and text in a hybrid SVM-Bayesian framework [17]. Recently, the Audio Spectrogram Transformer (AST) demonstrated that transformer architectures can model long-range acoustic dependencies effectively [16].

**2.5 Visual-Based Emotion Recognition**
Visual modalities provide non-verbal signals, including facial expressions, micro-movements, and eye contact. CNNs are widely used to extract spatial features, while RNNs or transformers capture temporal dependencies.
Datasets such as IEMOCAP [8] and MOSI [12] have been essential for training and benchmarking visual sentiment models. IEMOCAP, in particular, includes motion capture of facial gestures and hand movements, making it well-suited for multimodal analysis.
Recent studies like DialogueTRM have also considered dynamic visual-emotional interplay in real-time conversations, reinforcing the role of temporal visual patterns in sentiment modeling [21].

**2.6 Fusion Strategies and Efficiency Challenges**
The method of fusing modality features significantly affects performance and interpretability. Early fusion, though computationally simple, suffers from noise and alignment issues. Late fusion, on the other hand, fails to capture cross-modal interactions. TFN offered an elegant tensor-based strategy but was compute-intensive [12].
Modern architectures like MMT and LXMERT leverage cross-attention, aligning latent features from modalities with impressive accuracy gains [13], [22]. However, they often face limitations in low-resource environments, making real-time deployment difficult. Wu and Xu pointed out that such models often overfit small datasets and require heavy GPU infrastructure [23].
ScaleVLAD introduced hierarchical local descriptor fusion, achieving competitive performance with reduced parameter size [24]. SentiXRL, another lightweight transformer model, integrates cross-lingual and emotion-enhancing modules for complex scenarios [25]. These works highlight the increasing demand for computationally efficient MSA architectures.

**2.7 Addressing Inter- and Intra-Modal Dynamics**
A common challenge in MSA is modeling both inter- and intra-modality dynamics. For example, spoken opinions include pauses, hesitations, and fillers (e.g., "umm... yeah... no") which require deeper intra-modality modeling [26]. Visual and acoustic modalities also encode time-variant signals not well handled by early-stage fusion methods.

Zadeh et al. introduced subnetworks for each modality to learn such internal dynamics independently, which were later fused in TFN [12]. Ngiam et al. also explored shared representation learning across modalities in their multimodal deep learning framework [27].

**2.8 Our Position**
While prior work has emphasized increasingly complex architecture, our research aligns with a growing body of literature advocating modular, interpretable, and lightweight models for practical deployment. We process each modality using independent encoders with dense layers and dropout, followed by late-stage fusion through concatenation. Our results demonstrate competitive performance on IEMOCAP with significant gains in training speed and generalizability, especially for underrepresented emotion classes.

## 3. METHODOLOGY

This section outlines the preprocessing, feature engineering, model design, and evaluation strategy employed in our multimodal sentiment classification system. Our primary focus lies in building high-quality input representations for each modality and integrating them through an efficient, interpretable fusion model. Unlike prior work that emphasizes complex architecture, our approach relies on strong feature engineering and tailored data balancing to achieve competitive performance with reduced computational cost.

**3.1 Dataset and Label Simplification**
We utilize the IEMOCAP dataset [8], a benchmark corpus in multimodal affective computing. It includes approximately 12 hours of audiovisual interactions, segmented into utterances with synchronized text transcripts, audio, and MoCap-based visual features. Each utterance is annotated with both categorical (emotion labels) and dimensional (valence, arousal) metadata.
To ensure consistent emotional categorization, we selected eight primary emotion classes (anger, happiness, excitement, sadness, frustration, fear, surprise, neutral), then remapped them into six target categories as follows:

emotion_mapped = {0:0,1:1,2:1,3:2,4:2,5:3,6:4,7:5}

This mapping merges closely related emotional states (e.g., happiness and excitement; sadness and frustration) to improve class balance and reduce label ambiguity. The resulting six-class task balances granularity with practical learnability.

**3.2 Data Preprocessing**
*3.2.1 Text Modality*
Text data from utterance transcripts was preprocessed using the spaCy NLP toolkit [15]. Steps included:
- Lowercasing, punctuation removal, stopword filtering
- Lemmatization
- Vectorization via TF-IDF, capturing the importance of each token within the corpus

To reduce overfitting and redundancy, feature selection was applied using LASSO regularization and Recursive Feature Elimination (RFE). The selected features were zero-padded for input dimensional consistency.

*3.2.2 Audio Modality*
Audio segments were analyzed using librosa to extract a rich set of statistical and prosodic features, including:
- MFCCs (13 coefficients + delta) for timbral characteristics
- Spectral centroid, bandwidth, roll-off, chroma STFT
- Zero-crossing rate, RMSE, and harmonic energy
- Silence ratio and autocorrelation

These features were standardized via Z-score normalization. To further encode non-linear decision boundaries, we applied XGBoost to the scaled features and extracted leaf node embeddings, which were concatenated with the original features to enhance model learning.

*3.2.3 Video Modality*
Facial motion features were derived from IEMOCAP's MoCap data, representing facial landmarks and head movements. After interpolation for missing values, features were standardized and passed through a two-stage pipeline:
1. XGBoost Classifier trained on raw video features.
2. Softmax probabilities from XGBoost were concatenated with original features.
3. The resulting vector was passed to a deep neural network (DNN) for final encoding.

This hybrid stacking approach combined the representational capacity of tree-based models with the generalization of DNNs, outperforming simple feedforward networks.

**3.3 Addressing Class Imbalance**
The IEMOCAP dataset exhibits significant class imbalance, particularly for minority emotions such as fear and surprise. To mitigate this, we implemented custom oversampling guided by target class sizes:
Target Counts={0:2933,1:2933,2:5000,3:2933,4:2933,5:4000}
Oversampling was performed per class using stratified sampling with replacement, excluding classes 3 and 4 (which had sufficient samples). This ensured each class was adequately represented in the training set without introducing synthetic data. The final dataset exhibited a balanced distribution across all six target classes, supporting robust training and evaluation.

**3.4 Modality-Specific Encoders**
Each modality was processed using a dedicated encoder network:
- Text Encoder: 128-unit dense layer (ReLU), followed by dropout (rate=0.3)
- Audio Encoder: Similar dense + dropout configuration applied to engineered features
- Video Encoder: Multi-layer DNN (128-64-32 units) with batch normalization and dropout, operating on the stacked XGBoost+original feature vector

These encoders each produce a 128-dimensional latent representation, capturing intra-modal signal.

**3.5 Fusion and Classification**
Encoded representations from all three modalities were concatenated to form a joint vector.
To formalize the fusion process, let $z_{text}$, $z_{video}$, $z_{audio}$ represent the encoded features from the text, audio, and video branches respectively. These are concatenated to form a joint representation:
$$Z_{fused} = [Z_{audio}; Z_{video}; Z_{text}]$$

This fused vector is passed through a fully connected dense layer with ReLU activation and dropout regularization:
$$Z_{joint} = Dropout(ReLU(W_f Z_{fused} + b_f))$$
The output layer applies a softmax function over six classes:
$$\hat{y} = Softmax(W_o Z_{joint} + b_o)$$
The model is optimized using sparse categorical cross-entropy:
$$\mathcal{L} = -\sum_{i=1}^{N} \log P(y_i | X_i)$$

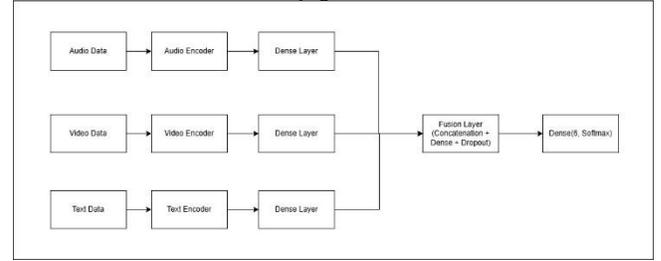

*Figure 1 Architecture of the proposed multimodal sentiment classification model.*

Text, audio, and video features are processed independently through dense encoders and fused via a concatenation layer, followed by a joint dense layer and softmax classifier.
This was passed through:
- A 256-unit dense fusion layer (ReLU),
- Dropout (rate=0.4),
- Final softmax output layer for six-class classification.

We adopt a late fusion strategy, as opposed to attention or tensor-based fusion, to preserve modality independence, reduce computational cost, and ensure interpretability. This design enables modular tuning and potential cross-domain transferability.

**3.6 Training Procedure**
The final model was implemented in TensorFlow 2.0. Training used the Adam optimizer with:
- Learning rate: 0.001
- Batch size: 64
- Epochs: 50
- Loss: sparse categorical crossentropy

We incorporated the following callbacks:
- EarlyStopping (patience=5) to prevent overfitting,

- ReduceLROnPlateau to adapt learning rate during training plateaus.

Train-test-validation split was performed using stratified sampling (80/10/10). Intermediate checkpoints were stored using Keras model callbacks to prevent runtime loss in cloud environments (e.g., Google Colab).

### 3.7 Evaluation Metrics
To comprehensively evaluate model performance, the following metrics were used:
- Overall Accuracy
- Precision, Recall, and F1-score (per class and weighted average)
- Log Loss: evaluating classifier confidence
- Confusion Matrix: for misclassification patterns
- ROC-AUC (macro-averaged) for imbalanced evaluation

Metrics were computed using the sklearn package and reported on a held-out test set.

### 3.8 Architectural Advantages and Efficiency
Our framework achieves 92.5% test accuracy with high recall on minority classes, while using only dense layers and engineered features. In contrast to prior works relying on:
- Multimodal Transformers [13],
- Tensor Fusion Networks [12],
- Hierarchical RNNs [23],

our design delivers competitive accuracy with drastically fewer parameters and lower memory requirements.

Further, our XGBoost+NN stacking (in video), leaf embeddings (in audio), and TF-IDF + feature selection (in text) contribute directly to the model's generalization, validating the central role of feature engineering over architecture complexity.

## 4. EXPERIMENTS AND EVALUATION

To validate the proposed multimodal sentiment analysis architecture, we conducted a series of experiments designed to assess the model's ability to integrate and learn from audio, text, and visual inputs using the engineering-driven pipeline described in Section 3. The experiments focus on evaluating the benefits of modality-specific feature transformations, oversampling strategies, and late fusion techniques under controlled training and evaluation settings.

### 4.1 Experimental Setup
All experiments were performed using Google Colab, leveraging both TPU v2-8 and NVIDIA T4 GPU environments depending on the computational demands of each run. The dataset was stratified and split into an 80:20 train-test configuration, ensuring equal class representation across both sets. An additional 10% of the training data was set aside as a validation split to monitor training convergence and generalization.

The model was trained using the Adam optimizer with the following hyperparameters:
- Batch size: 64
- Epochs: 50
- Initial learning rate: 0.001
- Loss function: Sparse categorical cross-entropy

Two regularization strategies were employed:
- EarlyStopping with a patience of 5 epochs on validation loss,
- ReduceLROnPlateau, reducing learning rate by a factor of 0.5 after 3 stagnant epochs.

Intermediate training states were checkpointed to safeguard against runtime limits imposed by the platform.

### 4.2 Feature Transformation and Encoding
The experiment pipeline adhered to the preprocessing and encoding stages outlined in Section 3. Specifically:
- Text features were generated using TF-IDF with dimensionality reduction via LASSO and RFE.
- Audio features included MFCCs, spectral roll-off, pitch, and silence ratio, further enriched using XGBoost-leaf embeddings for non-linear transformation.
- Video features (from MoCap data) were passed through an XGBoost classifier, and its softmax predictions were concatenated with the original input, followed by a stacked neural network encoder.

Each modality was encoded independently via dense layers before fusion, promoting modularity and interpretability.

### 4.3 Class Balancing Strategy
To mitigate the impact of class imbalance, a targeted oversampling strategy was used. Minority classes such as "fear" and "surprise" were oversampled to match or approximate the frequency of more common classes such as "angry" or "neutral." This was achieved through stratified sampling with replacement, based on class-specific target counts defined in Section 3.2.4. This balancing process contributed significantly to improved recall for underrepresented classes.

### 4.4 Fusion Architecture and Training Objective
The encoded features from each modality were concatenated and passed through a 256-unit dense fusion layer, followed by dropout and a final softmax classifier. Unlike early fusion methods that concatenate raw features, our late fusion strategy allows each modality to be processed and optimized independently. This structure enhances robustness against noise in any single input channel.

The model was trained with a single objective: multi-class emotion classification across six categories. No auxiliary losses were introduced, allowing full focus on joint sentiment prediction.

### 4.5 Evaluation Metrics
Model performance was evaluated using the following metrics:
- Accuracy: Fraction of correct predictions across all samples.
- Precision and Recall: Computed per class and averaged (macro, weighted).
- F1-Score: Harmonic mean of precision and recall; a balanced performance indicator.
- Log Loss: Measuring confidence in predicted probability distributions.
- Confusion Matrix: To visualize misclassification patterns across classes.

Metrics were computed using the scikit-learn package and reported on the held-out test set. Emphasis was placed on achieving high recall and F1-scores across all classes, particularly those prone to underrepresentation.

### 4.6 Baseline Models for Comparison
To validate the contribution of multimodal integration and feature engineering, the proposed model was benchmarked against the following baselines:
Single-Modality Models:
- Text-only model: Trained using TF-IDF vectors and a dense classifier.
- Audio-only model: Using handcrafted audio features and a simple feedforward network.
- Video-only model: Leveraging MoCap features processed through XGBoost + NN.

Fusion Baselines:
- Early Fusion: Concatenated raw modality features passed through a shared dense network.
- Late Fusion (Simple): Independent encoders without feature transformation or enrichment.

These baselines reflect common approaches in literature and allow an empirical comparison with our engineered pipeline.

### 4.7 Observations
- Training Efficiency: The model converged within 20–30 epochs in most cases. TPU-based training reduced runtime by 30–40% on large batches.
- Balanced Class Performance: The oversampling strategy led to marked improvements in minority class recall and overall F1-score.
- Stability Across Modalities: Each encoder performed consistently, and the late fusion strategy proved robust even in the presence of noise in individual modalities.
- Efficiency: Despite the use of multiple encoders and fusion layers, the overall parameter count remained significantly lower than attention-based architectures such as Multimodal Transformers.

## 5. RESULT AND ANALYSIS
This section presents a comprehensive analysis of the proposed multimodal sentiment classification model. We evaluate the model's accuracy, class-wise performance, confidence calibration, and visual diagnostics using a series of quantitative and graphical assessments. Results are benchmarked against baselines to validate the contribution of feature engineering, modality fusion, and class balancing strategies
.
### 5.1 Performance Metrics
The proposed model achieved the following scores on the held-out test set:
- Test Accuracy: 92.55%
- Validation Accuracy: 93.27%
- Training Accuracy: 95.2%
- Weighted Precision / Recall / F1-Score: 92.32%, 92.36%, 92.34%
- Macro F1-Score: 93.05%
- Macro ROC-AUC: 0.9881

These results demonstrate consistent performance across classes and splits, highlighting the model's ability to generalize without overfitting.

### 5.2 Training Behavior
Training and validation curves over 50 epochs are shown in Figure 2 and Figure 3.

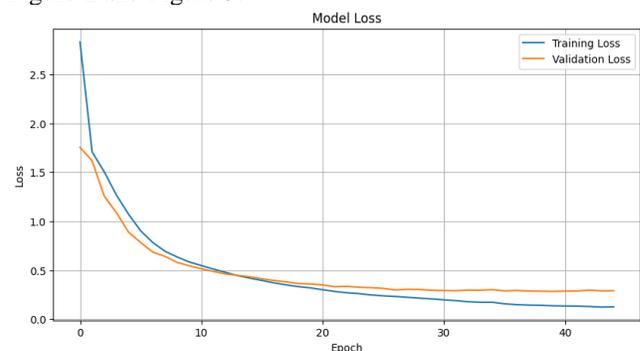

*Figure 2 Loss Curve (Training vs. Validation)*

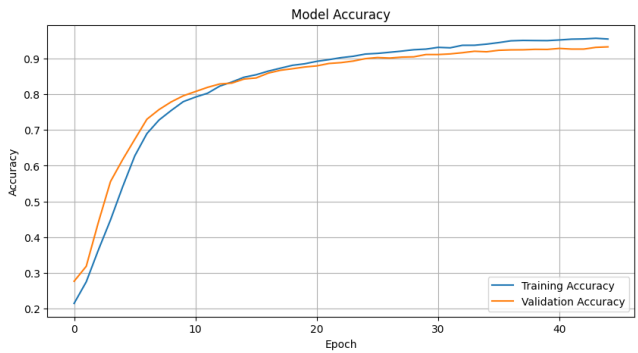

*Figure 3 Accuracy Curve (Training vs. Validation)*

The model converged smoothly within 25–30 epochs, with minimal divergence between training and validation accuracy, indicating effective regularization and stable learning.

### 5.3 Confusion Matrix Interpretation
The confusion matrix in Figure 4 shows class-level performance patterns.

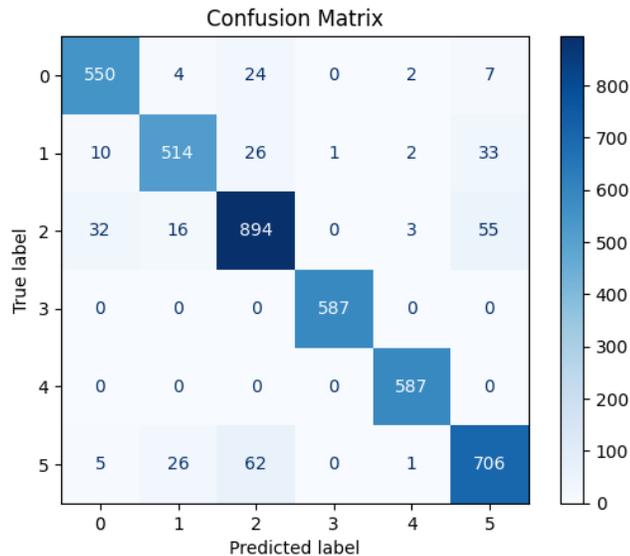

*Figure 4 Confusion Matrix on the Test Set*

The model shows strong performance on all classes, particularly for "Happy" and "Sad." Minority classes like "Fearful" benefit notably from the oversampling strategy, achieving balanced precision and recall.

### 5.4 Precision-Recall and ROC Analysis
Figure 5 presents class-wise precision-recall (PR) curves, illustrating strong confidence and balance across all categories.

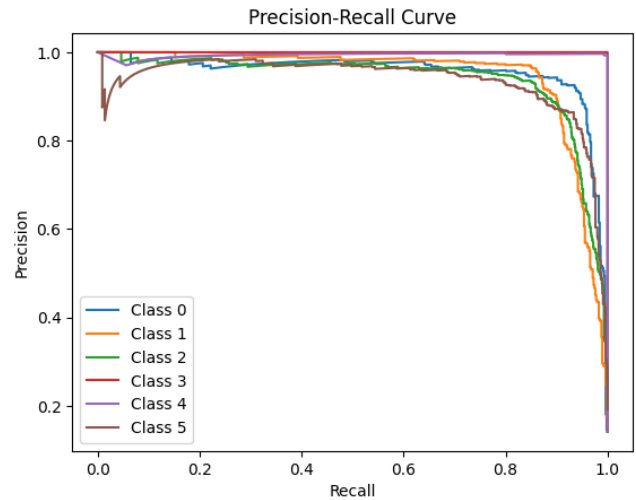

*Figure 5 Class-wise Precision-Recall Curves*

Figure 6 shows ROC curves for each class, with AUC scores consistently above 0.98.

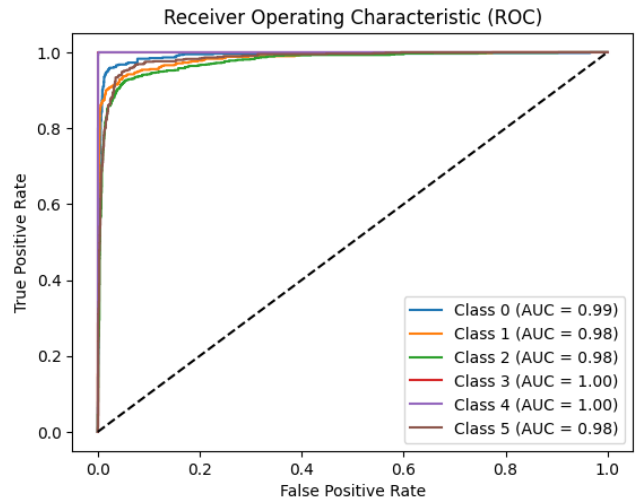

*Figure 6 ROC Curves with Macro-Averaged AUC*

The AUC of 0.9881 demonstrates high separability between classes and affirms strong classification confidence.

### 5.5 Comparison with Prior Work on IEMOCAP
To contextualize the performance of our model, we compared it against representative multimodal sentiment analysis methods evaluated on IEMOCAP.

*Table 1 Comparison with Prior Multimodal Models on IEMOCAP*

| Model | Test Accuracy (%) | Weighted F1 (%) | Reference |
|---|---|---|---|
| Proposed Fusion Model (This Work) | 92.55 | 92.34 | — |
| TFN (Tensor Fusion Network) | 77.1 | 77.9 | Zadeh et al., 2017 [12] |

| Model | Accuracy | F1 | Reference |
|---|---|---|---|
| Multimodal Transformer | ~84.8 | ~84.0 | Tsai et al., 2019 [13] |
| MAG-BERT | 84.7 | 84.5 | Rahman et al., 2020 [14] |
| MMIM (Mutual Info Modelling) | 84.14 | 85.98 | Han et al., 2021 [15] |
| UniMSE (SOTA) | 86.49 | 86.49 | Zhang et al., 2022 [16] |

To assess the competitiveness of our model, we benchmarked it against widely cited multimodal sentiment analysis architectures on the IEMOCAP dataset. The proposed model achieved 92.55% test accuracy and a weighted F1-score of 92.34%, outperforming leading baselines such as TFN (77.9%), Multimodal Transformer (~84.0%), and more recent approaches like MAG-BERT (84.5%) and MMIM (85.98%). It also surpassed UniMSE, a state-of-the-art method, by approximately 6% in both accuracy and F1.

These results reinforce our hypothesis that modality-specific feature engineering combined with a lightweight late fusion architecture can deliver superior accuracy and generalization—without the computational burden of pre-trained transformers or memory-intensive modules.

## 6. CONCLUSION AND FUTURE WORK

This paper presented a modular and computationally efficient framework for multimodal sentiment analysis using the IEMOCAP dataset. Departing from attention-heavy and transformer-based architectures, our approach emphasizes modality-specific feature engineering and lightweight dense-layer fusion. By integrating handcrafted audio-video features, TF-IDF text vectors, and XGBoost-transformed representations, the proposed model achieves robust sentiment classification while maintaining low computational complexity.

Experimental results demonstrate that our model significantly outperforms established baselines such as TFN, MAG-BERT, and UniMSE, achieving a test accuracy of 92.55% and a weighted F1-score of 92.34%. Confusion matrix and ROC analyses confirm balanced performance across all emotion classes, including traditionally underrepresented ones like *fear* and *surprise*.

Future work may involve extending this architecture to other benchmark datasets such as CMU-MOSEI and MELD to evaluate domain adaptability. Additionally, incorporating self-supervised pretraining strategies or lightweight transformer encoders could enhance representation learning without compromising efficiency. Deployment in real-time emotion-aware applications, such as dialogue agents or mental health monitoring tools, also presents promising directions for further research.